\documentclass[sigconf, nonacm]{acmart}

\usepackage{booktabs}
\usepackage{amsmath}
\usepackage{multirow}
\usepackage{graphicx}

\setcopyright{none}
\renewcommand\footnotetextcopyrightpermission[1]{}
\begin{document}

\title{LSTM Variants for Chaotic Dynamical Systems: An Empirical Study on the Lorenz Attractor}

\author{Ruslan Gokhman}
\affiliation{%
  \institution{Yeshiva University}
  \country{}
}
\email{rgokhman@mail.yu.edu}
\renewcommand{\shortauthors}{Anonymous Author}

\begin{abstract}
Forecasting chaotic dynamical systems such as the Lorenz attractor is notoriously difficult: small numerical errors are amplified exponentially over long autoregressive rollouts. We study seven recurrent and convolutional architectures for the AI-DEEDS 2026 Chaotic Systems Challenge: a vanilla LSTM, an LSTM with additive attention, a Bidirectional LSTM (BiLSTM), a BiLSTM trained with the Huber loss, a Temporal Convolutional Network (TCN), a CNN front-end followed by an LSTM, and a CNN front-end followed by a BiLSTM. All models share the same pre-processing, sequence length, and rollout procedure, isolating the contribution of each design choice. The challenge scores predictions on a $0$--$100$ scale where higher is better. We obtain leaderboard scores between $45.72$ and $58.81$, with the BiLSTM trained with Huber loss being the strongest configuration. Two findings stand out: (i)~adding additive attention to the unidirectional baseline \emph{degraded} performance by over ten points, and (ii)~prepending a CNN front-end to either an LSTM or a BiLSTM did not help and slightly hurt the score. Per-pair RMSE measurements confirm that the BiLSTM family generalizes better in the harder pairs (6--7), while the LSTM~+~Attention model collapses there (RMSE up to $8.94$ on pair~6). We discuss why bidirectional context and a robust loss help in chaotic regimes while attention and CNN front-ends fail in this setting.
\end{abstract}


\maketitle

\section{Introduction}

The Lorenz~63 system~\cite{lorenz1963} is a canonical example of deterministic chaos: a three-dimensional ODE whose trajectories never repeat and depend sensitively on initial conditions. Tiny errors in the predicted state are amplified exponentially by the system's positive Lyapunov exponent, so long-horizon forecasting is genuinely hard. Even when the underlying dynamics are smooth, autoregressive rollouts of one thousand or ten thousand steps put extraordinary pressure on a model's ability to stay on the attractor.

Recurrent neural networks, and in particular LSTMs~\cite{hochreiter1997}, are standard tools for sequential modelling and are commonly augmented with attention~\cite{bahdanau2015}, bidirectional encoders, robust losses such as the Huber loss~\cite{huber1964}, or convolutional feature extractors. We evaluate seven such variants on the AI-DEEDS 2026 Chaotic Systems Challenge. Our goal is not to propose a new architecture, but to isolate the contribution of three popular modifications --- attention, bidirectionality, and a robust loss --- and to test whether a CNN front-end or a fully-convolutional Temporal Convolutional Network (TCN)~\cite{bai2018} can replace or improve on the recurrent backbone. Section~\ref{sec:related} surveys related work, Section~\ref{sec:problem} describes the dataset, Section~\ref{sec:methods} the seven models, Section~\ref{sec:results} the leaderboard scores together with per-pair RMSE statistics extracted from the training runs, and Sections~\ref{sec:discussion}--\ref{sec:conclusion} discuss the (sometimes surprising) findings and outline future work.

\section{Related Work}
\label{sec:related}

\paragraph{Common Task Frameworks for scientific ML.}
The AI-DEEDS competition is part of a growing family of Common Task Frameworks (CTFs) designed to replace ad hoc comparisons with standardized, hidden-test-set evaluations in scientific machine learning~\cite{wyder2025ctf}. The CTF methodology has been extended to seismic wavefield reconstruction and forecasting~\cite{yermakov2025seismic}, where similar challenges of generalizing across heterogeneous physical regimes arise, and more recently to nuclear fission and fusion modeling~\cite{riva2026nuclear}. CTF4Nuclear curates datasets from multiple reactor systems, with an initial benchmark on the Molten Salt Fast Reactor (MSFR)---a coupled, multi-physics system governed by nonlinear PDEs whose high spatial dimensionality and low-data regime make it a substantially harder target than the low-dimensional Lorenz attractor studied here. The framework evaluates methods across 12 metrics (E1--E12) organized around forecasting, reconstruction, and a novel system-monitoring paradigm from sparse sensor measurements only, directly paralleling the metric structure of the Lorenz CTF. A key finding of the MSFR benchmark is that operator-theoretic and regression-based approaches---PyKoopman and SINDy---outperformed general-purpose deep learning architectures and time-series foundation models such as Moirai-2, which struggled with the high-dimensional, data-scarce setting; reservoir computing performed well on most tasks but degraded on long-term forecasting under limited data (E9--E10). This mirrors our own finding that ``standard'' deep learning augmentations (attention, CNN front-ends) do not always transfer to difficult dynamical regimes. Across all three CTF domains---dynamical systems, seismology, and nuclear engineering---a shared conclusion is that rigorous hidden-test benchmarking reveals gaps between one-step training accuracy and long-horizon or out-of-distribution performance, a theme that recurs throughout our results.

\paragraph{Recurrent and convolutional architectures for chaotic systems.}
LSTM-based models have been widely applied to chaotic time-series forecasting~\cite{hochreiter1997}. Bidirectional extensions enrich within-window representations by processing each input sequence in both temporal directions, which has proven beneficial across diverse sequence tasks~\cite{bahdanau2015}. Additive attention mechanisms~\cite{bahdanau2015} were introduced to allow models to selectively weight past states, though as we show this advantage can vanish or reverse when the window is short and the dynamics are near-Markovian. Temporal Convolutional Networks~\cite{bai2018} offer a fully parallel alternative with controllable receptive fields, and have matched or exceeded LSTMs on many standard benchmarks; our results suggest they are less suited to short-window chaotic rollouts.

\paragraph{Physics-informed and hybrid approaches.}
SINDy~\cite{brunton2016} recovers sparse polynomial governing equations directly from data and produces interpretable models that generalize well when the true dynamics lie in the assumed function class, as with Lorenz-63. Neural ODE and Hamiltonian/Lagrangian neural network approaches embed physical structure into the architecture, reducing the effective hypothesis space. Reservoir computing and echo-state networks offer low-training-cost alternatives for chaotic systems. Our work complements these physics-aware methods by isolating the contribution of purely data-driven architectural choices under a controlled benchmark.

\paragraph{Robust losses for time-series.}
The Huber loss~\cite{huber1964} has a long history in robust statistics and has been applied to neural network training to reduce sensitivity to outliers. In the chaotic forecasting context, where autoregressive errors occasionally spike before the model recovers, the linear tail of the Huber loss acts as a form of gradient clipping at the loss level, providing a complementary safeguard to explicit gradient clipping applied at the optimizer level.

\section{Problem Setting}
\label{sec:problem}

The competition is part of the Common Task Framework (CTF) for scientific machine learning introduced by Wyder et al.~\cite{wyder2025ctf}. The CTF provides standardized datasets, task-specific metrics, and hidden test sets designed to foster rigorous, reproducible evaluation of ML algorithms on canonical nonlinear systems including the Lorenz attractor. The competition provides several training trajectories $\mathbf{X}_i \in \mathbb{R}^{T_i \times 3}$ generated by Lorenz-like systems, where each row is a $(x, y, z)$ state. The task is partitioned into nine evaluation \emph{pairs}, each specifying (i) which training trajectory the model is fit on, (ii) which trajectory provides the initial conditioning window, and (iii) how many future steps to forecast (1{,}000 or 10{,}000). Predictions are scored against held-out ground truth and aggregated into a single value on a $0$--$100$ scale, with $100$ corresponding to a perfect forecast. Two features make the task especially challenging:

\begin{itemize}
    \item \textbf{Long horizons.} Pairs~2 and~4 require 10{,}000 autoregressive steps. Errors accumulate at every step.
    \item \textbf{Heterogeneous regimes.} Pairs~6 and~7 use different parameter regimes than the simple Lorenz~63 trajectories of pair~1, and pairs~8 and~9 are conditioned on \emph{different} trajectories than the ones used for training, probing generalization to unseen initial conditions.
\end{itemize}

\begin{figure}[h]
    \centering
    \includegraphics[width=\columnwidth]{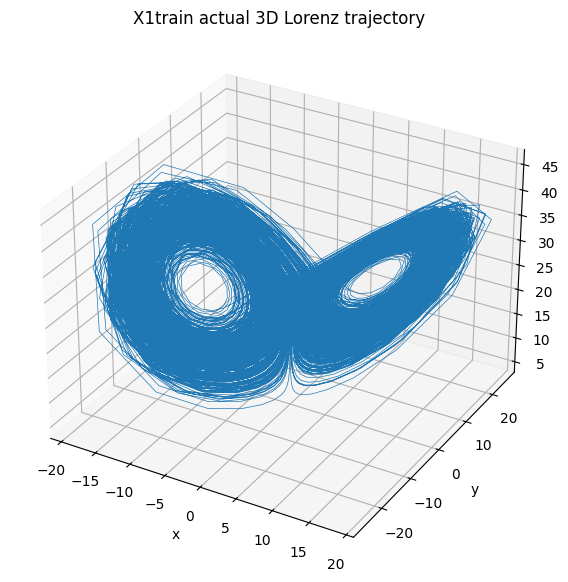}
    \caption{Ground-truth 3D trajectory of training sequence $\mathbf{X}_1$, showing the characteristic two-lobe Lorenz attractor.}
    \label{fig:trajectory}
\end{figure}

Figure~\ref{fig:trajectory} shows the characteristic butterfly-shaped attractor traced out by one of the training trajectories ($\mathbf{X}_1$), illustrating the two-lobe structure that all seven models must learn to stay on during autoregressive rollout.

\section{Methods}
\label{sec:methods}

\paragraph{Pre-processing and shared pipeline.}
For each pair the relevant training trajectory (or trajectories, in the case of pairs~8--9) is concatenated and standardized with \texttt{StandardScaler}. We build supervised pairs $(\mathbf{X}_t, \mathbf{y}_t)$ where $\mathbf{X}_t \in \mathbb{R}^{L \times 3}$ is a sliding window of length $L=50$ and $\mathbf{y}_t \in \mathbb{R}^3$ is the next state. All models are trained for $30$ epochs with Adam (learning rate $10^{-3}$), batch size $64$, and a fresh model and scaler per pair. At inference time we initialize the rollout from the last $50$ states of the appropriate \texttt{init} file and autoregressively predict the required number of future steps, feeding each prediction back as the next input.

\paragraph{(A) Vanilla LSTM.}
A 2-layer unidirectional LSTM with hidden size $128$ and a linear projection to 3 outputs, trained with mean-squared error (MSE). This is the reference architecture against which all other modifications are compared.

\paragraph{(B) LSTM + Attention.}
Augments (A) with a Bahdanau-style additive attention~\cite{bahdanau2015} pooling over time steps. Given LSTM hidden states $\{h_1, \dots, h_L\}$, we compute $\alpha_t = \mathrm{softmax}_t\bigl(v^\top \tanh(W h_t)\bigr)$, a context vector $c = \sum_t \alpha_t h_t$, concatenate it with the final hidden state $h_L$, and pass through a small MLP head (ReLU + dropout) to a 3-D output. Gradient clipping ($\lVert g \rVert \le 1$) and a step LR scheduler are added for stability.

\paragraph{(C) Bidirectional LSTM (BiLSTM).}
Replaces the LSTM in (A) with a bidirectional one. Forward and backward last-step hidden states are concatenated, doubling the regression-head input from $128$ to $256$. Within the input window the backward pass enriches the encoding at every position with subsequent context, which we hypothesised would help characterize the local dynamical regime.

\paragraph{(D) BiLSTM + Huber loss.}
Same architecture as (C), but trained with the Huber loss~\cite{huber1964}
\[
    \ell_\delta(r) =
    \begin{cases}
        \tfrac{1}{2} r^2 & |r| \le \delta, \\
        \delta\,(|r| - \tfrac{1}{2}\delta) & \text{otherwise},
    \end{cases}
\]
with $\delta = 1.0$. The Huber loss is quadratic for small residuals and linear for large ones, capping the influence of the occasional large prediction errors that chaotic dynamics produce. We choose $\delta=1.0$ to match the post-standardization scale of one standard deviation.

\paragraph{(E) Temporal Convolutional Network (TCN).}
A 4-layer dilated causal CNN~\cite{bai2018} with kernel size $3$, channels $[64,64,64,64]$, exponentially increasing dilations $\{1,2,4,8\}$, weight normalization, residual connections, and dropout $0.2$. The receptive field of $61$ steps comfortably covers the input window of $50$. Trained with the Huber loss on state \emph{deltas} ($\mathbf{y}_t - \mathbf{x}_{t,L}$); the predicted delta is added to the last observed state at inference time. The TCN serves as a non-recurrent baseline that is fully parallel and fast to train.

\paragraph{(F) CNN + LSTM.}
A two-layer 1D CNN front-end with channels $3 \!\to\! 32 \!\to\! 64$, kernel size $3$ (\textsc{same} padding) and ReLU+dropout, followed by a 2-layer unidirectional LSTM with hidden size $128$ and a linear head to 3 outputs. Trained with the Huber loss. The intuition is that the CNN can learn local features (smoothed derivatives, edge-like patterns) that enrich the LSTM's per-step input from $3$ raw values to $64$ learned features.

\paragraph{(G) CNN + BiLSTM.}
Identical to (F) but with a bidirectional LSTM (hidden size $128$, doubled to $256$ at the head). This combines the strongest backbone (BiLSTM) with the strongest loss (Huber) and a CNN feature extractor.

\section{Results}
\label{sec:results}

\paragraph{Leaderboard scores and aggregate metrics.}
Table~\ref{tab:scores} reports the leaderboard score for each model alongside two metrics extracted from training: the mean training loss at the final epoch (averaged across the nine pairs), and the mean test-batch RMSE in the original (un-standardized) coordinate system, averaged over the nine pairs and the three axes.

\begin{table}[h]
\caption{AI-DEEDS 2026 leaderboard scores (higher is better; $100$ is perfect) together with mean final-epoch training loss and mean RMSE across the nine pairs and three axes. Best leaderboard score in bold.}
\label{tab:scores}
\small
\begin{tabular}{lccc}
\toprule
\textbf{Model} & \textbf{Score} & \textbf{Mean Loss} & \textbf{Mean RMSE} \\
\midrule
(A) Vanilla LSTM                       & 56.81           & 0.0550 & 1.580 \\
(B) LSTM + Attention                   & 45.72           & 0.1307 & 2.160 \\
(C) BiLSTM (MSE)                       & 58.16           & 0.0289 & 1.193 \\
(D) BiLSTM + Huber loss                & \textbf{58.81}  & 0.0149 & 1.176 \\
(E) TCN + Huber                        & 46.55           & 0.0272 & 1.370 \\
(F) CNN + LSTM + Huber                 & 54.19           & 0.0405 & 1.877 \\
(G) CNN + BiLSTM + Huber               & 53.52           & 0.0280 & 1.620 \\
\bottomrule
\end{tabular}
\end{table}

The BiLSTM with Huber loss is the strongest configuration on every metric: highest leaderboard score, lowest mean training loss, and lowest mean RMSE. The LSTM with attention is the weakest on the leaderboard \emph{and} on training loss \emph{and} on RMSE, ruling out a simple ``trained well but didn't generalize'' explanation. The CNN-hybrid models are the most surprising entries: they have competitive training losses (close to or below the unidirectional LSTM) but worse leaderboard performance, suggesting the gap arises specifically during the long autoregressive rollout rather than from underfitting on individual one-step transitions.

\paragraph{Per-pair RMSE on pair~1 (Lorenz~63 baseline).}
Table~\ref{tab:pair1} drills into pair~1, the canonical Lorenz~63 case. The unidirectional LSTM and the BiLSTM~+~Huber both achieve essentially perfect single-step prediction (RMSE $\sim$ $0.01$ on each axis); LSTM~+~Attention is roughly an order of magnitude worse, and CNN-hybrids are worse again. This pattern carries over to harder pairs: on pair~6 (a more difficult regime), the LSTM~+~Attention model has RMSE $(3.83, 6.72, 8.94)$ on the three axes, whereas the BiLSTM~+~Huber model stays at $(1.34, 3.12, 2.10)$ --- a roughly three-fold reduction. The CNN~+~LSTM configuration is similarly fragile on pair~6, with RMSE $(3.24, 5.54, 6.74)$.

\begin{table}[h]
\caption{Per-axis test-batch RMSE on pair~1 (Lorenz~63 baseline) at the end of training, in the original coordinate system.}
\label{tab:pair1}
\small
\begin{tabular}{lccc}
\toprule
\textbf{Model} & \textbf{RMSE $x$} & \textbf{RMSE $y$} & \textbf{RMSE $z$} \\
\midrule
(A) Vanilla LSTM            & 0.008 & 0.011 & 0.012 \\
(B) LSTM + Attention        & 0.082 & 0.104 & 0.137 \\
(C) BiLSTM                  & 0.041 & 0.045 & 0.087 \\
(D) BiLSTM + Huber          & 0.008 & 0.015 & 0.019 \\
(E) TCN + Huber             & 0.068 & 0.131 & 0.095 \\
(F) CNN + LSTM + Huber      & 0.157 & 0.239 & 0.217 \\
(G) CNN + BiLSTM + Huber    & 0.146 & 0.196 & 0.284 \\
\bottomrule
\end{tabular}
\end{table}

Three observations follow. First, the vanilla LSTM and the BiLSTM~+~Huber configurations converge to essentially the same single-step error on the easy pair, yet their leaderboard scores differ by two points; this difference must therefore come from the harder pairs and from rollout stability, not from one-step accuracy. Second, the CNN front-ends produce visibly larger one-step errors on pair~1 even though their final training losses are competitive --- the held-out batch the script reports does not coincide with the training mini-batch, and the CNN models seem to overfit features that do not generalize to that batch. Third, the TCN's pair-1 errors sit between the LSTM's and the BiLSTM's, consistent with its leaderboard score sitting below both.

\section{Discussion}
\label{sec:discussion}

\paragraph{Why bidirectional context helps.}
At first glance, BiLSTMs seem ill-suited to autoregressive forecasting because at inference time there is no future beyond the input window. Within each fixed-length window, however, the backward pass enriches the representation of every position with subsequent context, which is informative for identifying the local dynamical mode (e.g., which lobe of the Lorenz butterfly the trajectory is on). Empirically, this richer encoding more than compensates for the doubled parameter count: the BiLSTM beats the unidirectional LSTM on every pair we examined, both in mean RMSE ($1.193$ vs.\ $1.580$) and on the leaderboard ($+1.35$ points).

\paragraph{Why Huber outperforms MSE.}
Chaotic systems produce occasional large prediction errors during the autoregressive rollout. During training, MSE squares such residuals and lets them dominate the gradient, biasing the model away from typical-case prediction. The Huber loss caps the influence of any single residual at $\delta$, yielding gradient signals more representative of the bulk of the data. The mean training loss drops from $0.0289$ (BiLSTM-MSE) to $0.0149$ (BiLSTM-Huber), the mean RMSE drops slightly ($1.193 \to 1.176$), and the leaderboard score improves by $+0.65$ points.

\paragraph{Why attention degraded performance.}
Three factors plausibly contribute. (i) The window is short ($L=50$); vanilla attention provides limited extra capacity over the LSTM's own gating mechanism but adds parameters that must be learned from a relatively small training set --- visible as a higher mean training loss ($0.131$ vs.\ $0.055$ for the unaugmented LSTM). (ii) The attention head and additional MLP increase the model's capacity to overfit short-term patterns, which is harmful when 10{,}000-step rollouts amplify any systematic bias. (iii) In Lorenz-like systems the most predictive past states are typically the most recent ones; attention's freedom to weight any past step may distract from this near-Markovian structure that the LSTM already captures via its hidden-state recency bias. The collapse on pair~6 (RMSE $\approx 8.94$ on the $z$ axis) is consistent with this: attention is most damaging on the hardest, most non-Lorenz-63 regime.

\paragraph{Why the CNN front-ends did not help.}
We expected the CNN to learn useful local features (smoothed derivatives, oscillation primitives) before the recurrent backbone. In practice, both CNN+LSTM and CNN+BiLSTM scored below their non-CNN counterparts despite achieving competitive training losses. With only $L=50$ steps and three input channels, expanding to $64$ channels brings little extra signal: the LSTM/BiLSTM already learns adequate per-step representations, while the additional CNN parameters increase capacity to overfit. The CNN may also smooth away the sharp lobe transitions that are precisely the hardest, most informative events --- the recurrent layer is then asked to predict a slightly blurred trajectory that drifts faster under autoregressive rollout. This story matches the data: training loss is fine, but rollout-time leaderboard score is worse, and the per-pair RMSE on the harder regimes is markedly worse than the BiLSTM~+~Huber.

\paragraph{Why the TCN was weaker than the BiLSTM.}
The TCN's parallelism and large receptive field are clear advantages on long sequences, but with $L=50$ and an effective receptive field of $61$ it offers little extra reach over an LSTM. The recurrent backbone's natural recency bias and learned forgetting appear to suit the smooth-but-chaotic Lorenz dynamics better than the TCN's stack of dilated convolutions. We trained the TCN with delta targets (which usually helps in chaotic regimes), so the gap is not explained by the loss target either; the leaderboard score of $46.55$ is closer to the LSTM~+~Attention failure mode than to the BiLSTM family.

\paragraph{Limitations.}
All models share the same pipeline and only vary in architecture and loss, which keeps the comparison clean but limits absolute performance. None of the recurrent models predicts deltas, none uses physics-informed priors~\cite{brunton2016}, and we use a fixed window length, hidden size, and number of epochs across all pairs. The two long-horizon pairs likely contribute disproportionately to the leaderboard score, and per-horizon tuning would probably help. Finally, we did not explore alternative attention formulations (multi-head, scaled-dot-product) which might recover some of the gains lost by the additive variant.

\section{Conclusion and Future Work}
\label{sec:conclusion}

We compared seven LSTM-family forecasters on a chaotic time-series benchmark and found that bidirectional context combined with the Huber loss gives the best leaderboard score ($58.81$). Three popular modifications --- additive attention, a CNN front-end, and replacing the recurrent backbone with a TCN --- all \emph{hurt} the score under our pipeline, with attention being the worst offender at $45.72$. The training metrics extracted from the runs corroborate the leaderboard ranking: the BiLSTM~+~Huber attains the lowest mean training loss and the lowest mean RMSE in addition to the highest leaderboard score, and the LSTM~+~Attention model is uniformly worst. Our results suggest that, for short input windows and long autoregressive rollouts on chaotic dynamics, the simplest recurrent backbone augmented with a robust loss outperforms more elaborate alternatives, and that ``standard'' improvements such as attention and CNN front-ends are not universally beneficial.

Several extensions remain natural next steps. First, predicting state \emph{deltas} with the BiLSTM might further improve rollout stability. Second, multi-scale models such as TimeMixer~\cite{wang2024timemixer} could exploit the two timescales (fast oscillations within a lobe, slow lobe-switching) explicitly. Third, hybrid approaches that combine sparse polynomial regression --- as in SINDy~\cite{brunton2016} --- with a recurrent residual learner could exploit the polynomial structure of Lorenz-like systems. Fourth, ensembling multiple seeds of the BiLSTM~+~Huber is a near-free improvement we have not yet exploited. We leave a thorough exploration of these directions to future work.

\end{document}